\documentclass[conference]{IEEEtran}
\IEEEoverridecommandlockouts

\usepackage[nottoc]{tocbibind}
\usepackage{amsmath,amssymb,amsfonts}
\usepackage{algorithmic}
\usepackage{enumitem}
\usepackage{graphicx}
\usepackage{tikz}
\usepackage{booktabs}
\usepackage{url}
\usepackage{multirow}
\usepackage{textcomp}
\usepackage[numbers]{natbib}
\usepackage{svg}
\usepackage{tabularx}
\usepackage{array}
\usepackage{xcolor}
\def\BibTeX{{\rm B\kern-.05em{\sc i\kern-.025em b}\kern-.08em
    T\kern-.1667em\lower.7ex\hbox{E}\kern-.125emX}}
\begin{document}
\newcolumntype{C}{>{\centering\arraybackslash} m{6cm} }
\title{Using Visual and Vehicular Sensors for Driver Behavior Analysis: A Survey
}

\author{\IEEEauthorblockN{Bikram Adhikari}
\IEEEauthorblockA{
\textit{George Mason University}\\
Fairfax, Virginia, USA \\
badhika5@gmu.edu}
}

\maketitle

\begin{abstract}
Risky drivers account for 70\% of fatal accidents in the United States. With recent advances in sensors and intelligent vehicular systems, there has been significant research on assessing driver behavior to improve driving experiences and road safety. This paper examines the various techniques used to analyze driver behavior using visual and vehicular data, providing an overview of the latest research in this field. The paper also discusses the challenges and open problems in the field and offers potential recommendations for future research. The survey concludes that integrating vision and vehicular information can significantly enhance the accuracy and effectiveness of driver behavior analysis, leading to improved safety measures and reduced traffic accidents.
\end{abstract}

\begin{IEEEkeywords}
CANBUS, Machine Learning, Deep Learning, Behavioral Analysis 
\end{IEEEkeywords}

\section{Introduction}
In recent years, with technological advancements and societal needs, there has been a lot of interest in analyzing human behavior and detecting human perception in real-time \citep{yu2020human}. Similarly, with the advancement of intelligent transportation systems, availability of data, and advancements in multiple sub-data sources, we observe an increasing interest in driver behavior analysis (DBA) and prediction for a better driver experience, road safety, and improved traffic flow \citep{9430766}. This is increasingly important as driver behavioral traits accounted for almost 70\% of the fatal road crashes in the USA in 2020 \citep{nhtsa_2022}.

According to \citep{s151229822}, driving behavior can be classified into one of the four major groups: Aggressive driving, Inattentive Driving, Drunk Driving, and Normal Driving. 

One of the common assessments for driver behavior comes from knowledge-based modules. These modules predict a result based on the current state of observation and subjective knowledge. As a subpart, the Threshold-based model \citep{8844776} operates by setting the thresholds for classification based on personal and experimental results. Though it performs well with binary classification, the threshold-based approaches suffer as we increase the number of classes. Also, this approach benefits from the increase in parameters, but this increase further complicates the system and has an undesirable computational overhead. Alternatively, insurance companies relied on fuzzy inferences for a score or continuous prediction of driver behavior and risk associated with driving. These modules proved beneficial over the threshold-based modules by providing classification on a continuous scale rather than a discrete assessment of driver behavior \citep{s151229822}. 

To standardize accuracy, assessment metrics such as PERCLOS (Percent of time Eyelids are CLOSed) \citep{wierwille1994research} are used to measure the cognitive vigilance of the driver while engaged in driving activities. It measures the percentage of time a person's eyes are closed at least 80\% over the pupil. Even to this day, most of the research involving visual inference of the driver \citep{8949469} relies on the PERCLOS measure for the assessment of driver behavior.

With advancements in sensor technologies and computational efficiency, the project's focus has shifted to machine learning-based assessment of driver behavior. However, the selection of machine learning tools relies on the available data sources. For DBA, the data sources can be divided into visual and non-visual, focused on either the driver or the vehicle and the surrounding environment. Biomedical sensors like Electroencephalography have shown the highest accuracy in the assessment of DBA, but these systems are intrusive, noisy, and pose a challenge for large-scale implementation. Visual and vehicular features provide scalable and efficient solutions but at the cost of the accuracy of the assessment.

This survey summarizes the work on Vision and Vehicular Information for DBA and the progressive advancements in providing comparable accuracy with the intrusive features of behavioral assessment. Section II discusses the preliminary requirements to follow through with the paper. Section III introduces the taxonomies of my analysis, which are followed by a detailed taxonomy-based literature review in Section IV. Section V wraps up the literature review with a discussion of open problems that may guide future research in this field, and the paper concludes with final remarks in Section VI.

\section{PRELIMINARIES}
 \begin{figure*}[h]
 \begin{center}
  \includegraphics[width=\linewidth]{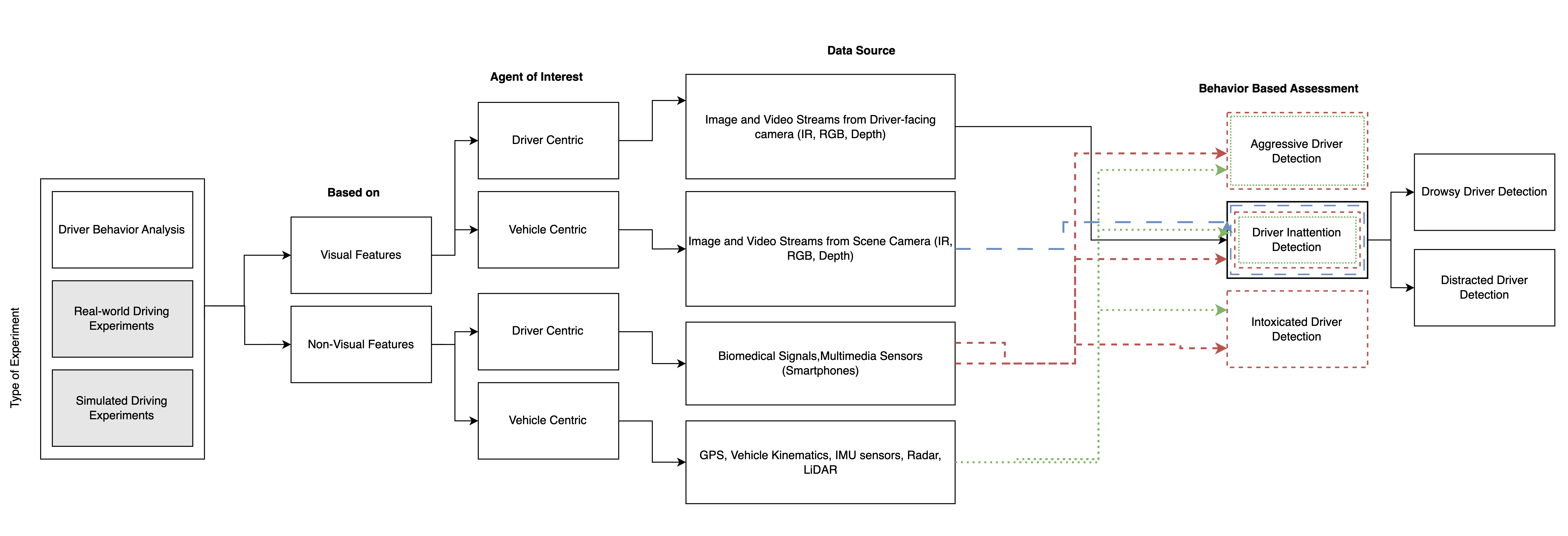}
  \caption{Taxonomical Illustration of Driver Behavior Analysis based on the Experiment type, data source, and the use case}
 \end{center}
\end{figure*} 
\subsection{Driver Behavior/Style classification}
Driver behavior refers to the subconscious and conscious actions of a driver while interacting with the vehicle, the road environment, and other vehicles in the environment. Traditionally, researchers \citep{s151229822} have defined safe driving behavior as the expected behavior of drivers in their day-to-day loco-motion, whereas the driving behavior of a specific driver under the influence of physical or mental stress is categorized as abnormal driving behavior. However, such a one-class classification of driving behavior seems oversimplistic because driving patterns vary from person to person, and a certain pattern that could be considered abnormal for an individual could be a typical driving trait for another individual. Therefore, in their survey, \citep{s151229822} classify driver behaviors into four major categories:
\subsubsection{Aggressive Driving style}
		An aggressive driving style is an intentional behavioral pattern of a driver associated with high-risk speeding profiles. The usual behavioral patterns observed include irregular, instantaneous, and abrupt changes in vehicle speed, improper lane management, and excessive or inconsistent acceleration and deceleration \citep{s151229822}. Such driving behaviors are typically observed due to drivers' impatience, annoyance, hostility, or attempts to minimize their commute time.
\subsubsection{Inattentive Driving style}
    Unlike the intentional and somewhat routine aggressive driving style, the inattentive driving style is identified by an instantaneous deviation from normal behavior followed by a sudden correction from the driver to return to normal driving conditions \citep{s151229822}. According to the "100-Car Naturalistic Study" \citep{neale2005overview}, inattentive driving is found to be the cause of 78\% of total accidents and 65\% of near accidents recorded. Driver distraction and driver fatigue (drowsiness) are considered to be the major causes of inattentive driving style. 
\subsubsection{Intoxicated Driving style}
    Intoxicated Driving is attributed to driving under the influence of alcoholic and non-alcoholic drugs. It is categorized by deterioration in driving with a periodic trait of aggressive and inattentive driver behavior due to lack of self-discipline and concentration \citep{s151229822}. Being a prime cause of accidents, in-vehicle detection of Driving Under the Influence (DUI) is an active area of research in providing driver feedback and assistance in semi-automatic intelligent vehicles. 
\subsubsection{Safe/Normal Driving style}
   After categorizing different abnormalities in driving, the normal or safe driving style is defined as a behavioral pattern of a driver when they avoid the above-mentioned risk-inducing abnormalities \citep{s151229822}. This includes maintaining proper acceleration and deceleration, avoiding tailgating, and more. However, the assessment of normal driving style is a highly subjective analysis, and the norm of safe driving might differ from person to person and scenario to scenario. Therefore, the establishment of context-aware systems to analyze safe and aggressive driving styles is an active area of research, which will be discussed further in the upcoming sections of this survey. 
\subsection{Data sources for driver behavior analysis}
\subsubsection{Visual Sensors (Cameras)}
	Visual sensors are the most widely used source for DBA. Visual cues of the driver can help detect the driver’s state of attentiveness and cognitive distractions such as talking on the phone, texting while driving, and more. Studies such as \citep{8436988}-\citep{ghosh2021speak2label} analyze driver behavior with over 90\% accuracy by capturing the driver's eye movements, changes in facial expression and posture, head and body posture estimates, and by using state-of-the-art computer vision and deep learning tools. In addition, \citep{9618784} discusses in-vehicle camera modules used for detecting driver drowsiness and distraction, along with a list of publicly available datasets based on these methods. Visual sensors offer variations in the data collected using stereo cameras for depth perception, and IR cameras in case of low illuminance. Furthermore, visual sensors are highly cost-efficient, with a survey conducted by \citep{8844776} discussing all the research focused on the use of smartphone cameras for DBA. However, these sensors do suffer from issues such as illuminance, partial glances, and occlusion of the target space.
  
\subsubsection{Physiological Sensors (Electroencephalograms)}
There are physiological changes that individual projects while under stress, or fatigue and sleepiness \citep{Kaplan2015DriverBA}. Sensors like Electroencephalograms (EEG) monitor the alpha and beta waves from the brain, Electrocardiograms (ECG) monitor the heart rate, and Electrooculograms (EOG) monitor the abnormalities in the retina. As they have high accuracy and are successful in predetermination of driver stress if it is physically reflected as they are placed directly on the user's body \citep{Kaplan2015DriverBA}. However, because of their intrusive nature (attached to the driver), these sensors can themselves act as a distraction and cause deviation from normal driving behavior \citep{8844776}. The sensors are also highly susceptible to noise and movements which can reduce the efficiency of the analysis. As there are preexisting solutions that offer comparable accuracy and are less tedious to operate, these physiological sensors are not suitable for the practical implementation of DBA. Thus, this survey will not cover the DBA approaches based solely on the physiological sensors. However, the readers can refer to \citep{panicker2019survey} for a comprehensive review of various approaches to DBA using physiological sensors. 
\subsubsection{Multimedia and Portable sensors (Smartphone)}
	Smartphones offer a cheap, readily available, and easily adaptable means of data collection. Other multimedia tools that smartphones come with, like the microphone, can be used to detect different driver behaviors based on yawning, nodding, or using multimedia \citep{8844776}. Equipped with front and back cameras, microphones, accelerometers, gyroscopes, and GPS, modern smartphones can track various operational tasks, such as driver style assessment, distraction, and interaction with the external environment. However, due to cost constraints with smartphone devices, the sensors used are not as precise as the standalone sensors available for experiments. The data collection itself is highly susceptible to the orientation of the smartphone and is prone to have noisy and false readings, even with the slightest change in the device's orientation \citep{8844776}\citep{Kaplan2015DriverBA}.

\subsubsection{Vehicle Internal sensors (CANBUS)}
Introduced by Robert Bosch GmbH as a standardized communication protocol for vehicles in the mid-1980s \citep{8412228}, CANBus offers more than 2000 vehicular kinematic signals ranging from speed and lateral acceleration to the turn-signal light blink status \citep{azadani2021driving}. CANBus provides a non-intrusive, easily adaptable, and accessible data source in real time. The CANBus is also highly robust to the illuminance and occlusions that are prominent in the real driving scenario. However, CANBus encryption is vehicle and model-specific and also doesn't take into account the driver's experience, road, and traffic conditions while evaluating the driver's behavior \citep{s151229822}.

\subsubsection{Vehicle External Sensors (LiDAR, GPS, etc.)}
External sensors, such as GPS, provide indirect and precise measurements of a vehicle's speed, acceleration, and position, which are beneficial in mapping the driver's interaction with the external environment. More advanced sensors, such as LiDAR and Radar, also provide a robust and accurate perception of the vehicle's external environment. These data sources usually complement the information gathered from other sensors to make better predictions of driver behavior \citep{azadani2021driving}. However, incorporating these sensors can be difficult, and they are often expensive. Furthermore, they have pre-existing limitations in adverse weather and road conditions, such as GPS.

\subsection{Methods used for Driver Behavior Analysis}
\subsubsection{Knowledge Based Approach}
Initial studies of DBA relied on Knowledge-Based and threshold models for binary classification of driver behavior \citep{8844776}. One of the first standardized measures for the detection of driver drowsiness comes from PERCLOS, given as :

\begin{equation}
    \text{PERCLOS} = \dfrac{\text{Time when eyes are closed at least 80\%}}{(\text{Closed eyes time} + \text{Open eyes time})} X 100 \\
\end{equation}

Usually, the PERCLOS value is higher for the drowsy driver than for the awake driver as drowsy and tired drivers tend to have a higher eye closure duration. Instead of hard thresholds, fuzzy logic represents the degree of truth in any analysis \citep{s151229822}.

State machines are models that switch between different states based on the knowledge they receive. Deterministic Finite State Machines (FSM) have a single transition for any given input, unlike a non-deterministic FSM which can transition to multiple different states given the input in a particular state \citep{s151229822}. FSMs are typically used in Advanced Driver Assistance Systems (ADAS) where, based on the vehicular position and driver's performance, the ADAS sets a vehicle state and either corrects or alerts the driver to avoid any accidents. These systems don't require a large ground truth dataset after the initial rules have been established. However, they do suffer from generalizability as the thresholds for vehicles, humans, road, and sensor types differ from one to another \citep{8844776}.

Similar to FSM, Hidden Markov Models are used in research to model the temporal evaluation of the driver's state \citep{bhatt2016novel}. Research using HMM models uses vehicular parameters like acceleration, steering angle, and vehicle speed, and then identifies and analyzes the hidden state like driver distraction and attitude \citep{s151229822}.

\subsubsection{Mathematical Tools based approach}
Mathematical tools are useful for the classification and analysis of temporal data, specifically CANBus data of vehicle kinematics in the case of DBA \citep{s151229822}. Fast Fourier Transform (FFT) is used to analyze the frequency of occurrence of sudden corrections based on the assumption that a conscious driver performs gradual acceleration and steering wheel corrections, to detect distraction or drowsiness of the driver.

Dynamic Time Wrapping (DTW) has also been extensively used for temporal data-based DBA \citep{s151229822}. DTW classifies the driver's current behavior by computing the distance of the temporal sequence with pre-defined templates.

Kalman Filter is extensively used in temporal data sequences to filter out the noise in the data and estimate the value based on past events. DBA adopts Kalman filtering for gaze estimation, tracking the vehicular position, and prediction of future driving events for better assessment of the driver's behavior \citep{azadani2021driving}.

\subsubsection{Classical Machine Learning Models}
Knowledge-based DBA is limited to a specific vehicle model range and choice of threshold. Mathematical tools are generally employed with temporal signals. These tools perform well and are simple to implement, but they suffer from the choice of threshold or acceptance criteria and rely heavily on perfect and noise-free data \citep{azadani2021driving}. Both knowledge-based and mathematical tool-based models are sensitive to the number of parameters used for analysis, with systems performing better with more classification parameters. Recently, there has been a dominant shift towards the use of machine learning for DBA. A survey \citep{elamrani2020engineering} discussing the application of machine learning techniques for DBA illustrates that more than 72\% of DBA research has adopted machine learning techniques, specifically Support Vector Machines (SVM), Neural Networks (NN), Bayesian Learners (BL), and Ensemble learners (EL). Before the advancement in deep learning and neural networks, SVMs were considered the best model for higher-dimensional feature analysis like multi-sensor DBA. Using the kernel trick to operate in a higher dimensional space, SVMs provided a decent analysis of driver behavior without suffering from the curse of dimensionality \citep{tango2013real}. Bayesian algorithms are probabilistic methods of classification based on conditional dependence between random variables, using Bayes's theorem. Ensemble learners, like Random Forest, are a combination of multiple independently trained models. This approach allows multiple models to analyze and classify the data, but training and parameter tuning of multiple algorithms make it computationally expensive and inefficient in real-world implementation \citep{8844776}.
\subsubsection{Deep Learning Models}
Deep Learning models replicate the interaction of neurons (represented as nodes) in the nervous system to collectively analyze problems. With advancements in neural network models, from basic feed-forward single neurons to dense generative models, and vehicular digitization and implementation of ADAS systems, there has been a lot of work using DNNs in ADAS research. Even though these models require heavy computational resources, large data corpus, efficient preprocessing, and hyper-parameter tuning, they can extract hidden features embedded in the dataset that classical models fail to. Convolutional Neural Networks (CNNs) and Deep Neural Networks (DNNs) are specifically used for visual data processing and analysis, providing state-of-the-art performance in evaluation. Recurrent Neural Networks benefit temporal data analysis, with models like Long-Short Term Memory (LSTM) accessing the driver profile based on the temporal data with as high as 96\% accuracy \citep{8778416}.
 \begin{figure}[h]

    \centering
    \includegraphics[width=25em]{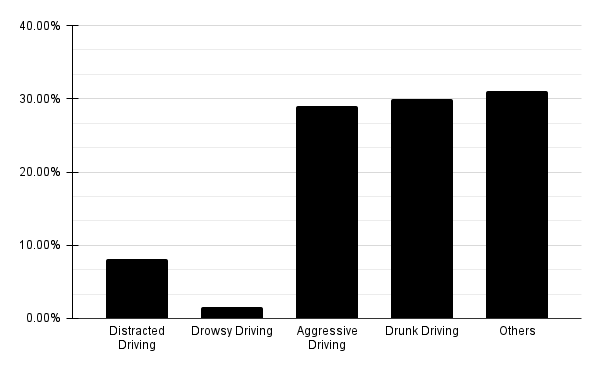}
    \caption{Percentage of traffic fatalities based on driver behavioral attributes-NHTSA 2020 [\citenum{nhtsa_2022}]}

  \label{fig:01}
\end{figure} 
\section{TAXONOMY}
Based on the data acquisition sources and the classification and analysis approaches mentioned before, driver behavior is primarily analyzed for:

\subsection{Aggressive Driver Detection}
NHTSA \citep{nhtsa_2022} reported 11,258 fatalities in 2020 caused by speeding, which accounts for 29\% of overall traffic fatalities, as mentioned in Figure \ref{fig:01}, and 2,564 fatalities in crashes involving hit-and-run. Washington D.C. alone reported 2,537 aggressive driving-related injuries between 2017-2021, attributing to 19.8\% of all driving-related injuries \citep{vision_zero_dc}. With a significant impact on individuals within the vehicle as well as other traffic and pedestrians, there have been significant advances in aggressive driver detection. The visual feature-based (camera-based) modules monitor the speed, lane positioning of the vehicle, and the distance between vehicles to detect, analyze, correct, and in some cases report aggressive driving behavior, whereas the telematics-based approach relies on vehicle-centric non-visual features for the assessment of driver aggressiveness.  
\subsection{Driver Inattention Detection}
As mentioned before, driver inattention can be caused either by distraction or fatigue (drowsiness). In 2020, distracted driving behavior contributed to 3,142 (8.1\%) reported fatalities, and drowsy driving behavior contributed to 633 (1.6\%) reported fatalities \citep{nhtsa_2022}. Although driver inattention seems to have lower rates of fatalities, these behavioral attributes contribute to more than 400,000 ($>$20\%) of overall driving-related injuries in the USA in 2020 \citep{nhtsa_2022_dis}. Unlike continuous behavioral patterns such as Aggressive and Drunk, accidents can be avoided in the case of driver inattention by providing real-time detection and feedback to the driver.

The driver-centric visual features rely on head and gaze estimation to predict driver inattention. Some research also analyzes facial features such as FOM (Fraction of Mouth-open) and other bodily features such as pose and hands on the steering wheel to detect driver inattention while driving. The most accurate assessment of driver inattention comes from biomedical sensors attached to the driver. Changes in heart rate and EEG signals can help detect driver inattention, especially drowsy drivers in the preliminary stages of inattention.

Vehicle-centric approaches rely on sensors and cameras to estimate driver inattention using the frequency of brake and throttle pressed, steering angle, lane markings, and position with respect to other vehicles.

\subsection{Intoxicated Driver Detection}
While it is illegal to drive under the influence of alcohol or drugs (above a certain level), it is alarming that driving under the influence of alcohol is responsible for 11,654 (30\%) of the total fatal traffic accidents in 2020 \citep{nhtsa_2022}. Breathalyzer tests are still the most efficient means of detecting alcohol influence. However, biomedical sensors such as heart-rate monitors have shown efficiency in detecting intoxicated drivers under the influence of both alcoholic and non-alcoholic substances. Driver-centric sensors have also shown promising results in detecting intoxicated drivers, using cameras for gaze and facial expression analysis, microphones for detecting slurred speech, and vehicular sensors to detect intoxication based on the strength of the steering wheel grip and lateral position of the vehicle. 

\begin{table*}[h]
  \centering
  \caption{Overview of the Data sources available for Driver Behavior Analysis}
  \label{tab:00}
  \begin{tabular}{|m{2mm}|m{2mm}|m{20mm}|m{40mm}|*2{m{45mm}|}}
\hline
\cline{3-6} 
&&Data Source& Data Features& Advantage & Disadvantage\\
\hline 
{{\rotatebox[origin=c]{90}{Visual Features}}} & {{\rotatebox[origin=c]{90}{Driver Centric}}} & Image and Video Streams from Driver-facing camera & 
\begin{itemize}[noitemsep,topsep=2pt,leftmargin=*]
      \item RGB and IR image and video sequence especially focused around the face, eye, and mouth tracking  
\end{itemize}
&
\begin{itemize}[noitemsep,topsep=2pt,leftmargin=*]
      \item Use of SOTA computer vision and visual processing techniques
      \item Robust, Non-intrusive and cost-adaptive
      \item Efficient in preliminary detection of distraction for accident avoidance. 
\end{itemize}
&\begin{itemize}[noitemsep,topsep=2pt,leftmargin=*]
      \item Privacy concerns with driver recordings, face, mouth, and pose analysis
      \item Sensitive to illumination, occlusions, and orientation
      \item Requires a large amount of labeled data for efficient analysis
      \item High computation time and resources. 
\end{itemize} \\
\cline{2-6}
& {{\rotatebox[origin=c]{90}{Vehicle Centric}}} & Image and Video Streams from Scene Camera & 
\begin{itemize}[noitemsep,topsep=2pt,leftmargin=*]
      \item RGB, Depth and thermal image and video sequence focused around the scene the driver is interacting with
\end{itemize}&
\begin{itemize}[noitemsep,topsep=2pt,leftmargin=*]
      \item Road condition analysis, markings, and traffic identification and depth and flow perception using SOTA Computer vision and deep learning techniques. 
      \item Robust, Non-intrusive and cost-adaptive
\end{itemize}
&
\begin{itemize}[noitemsep,topsep=2pt,leftmargin=*]
      \item Sensitive to weather, time of day, vehicle's position, and poor road and traffic conditions.
      \item Computationally expensive and require a large amount of data so problematic in real-time implementation. 
\end{itemize}
\\
\hline
{{\rotatebox[origin=c]{90}{Non-Visual Features}}}
&{{\rotatebox[origin=c]{90}{Driver Centric}}} & Biomedical Signals and Multimedia signals &
\begin{itemize}[noitemsep,topsep=2pt,leftmargin=*]
      \item EEG, EOG, ECG sensors for driver's physiological measurements
      \item Microphone, Gesture, and Reaction time sensors  
\end{itemize}
&
\begin{itemize}[noitemsep,topsep=2pt,leftmargin=*]
      \item Highly accurate 
      \item Anonymous identification 
\end{itemize}
&
\begin{itemize}[noitemsep,topsep=2pt,leftmargin=*]
      \item Intrusive and could be a form of distraction itself 
      \item Complicated and noisy data that requires a lot of post-processing
      \item Data source highly susceptible to movements and only applicable in controlled settings. 
\end{itemize}

\\
\cline{2-6}
& {{\rotatebox[origin=c]{90}{Vehicle Centric}}} & Vehicle Kinematics and External Environment Sensor Signals & \begin{itemize}[noitemsep,topsep=2pt,leftmargin=*]
      \item CanBUS data of throttle and brake pressure, steering angle, speed, lateral and longitudinal acceleration
      \item GPS, LiDAR, IMU,  RADAR sensors
      \item Smartphone-based Gyroscope, accelerometer sensors
\end{itemize}&
\begin{itemize}[noitemsep,topsep=2pt,leftmargin=*]
      \item Precise and accurate assessment of vehicular kinematic information, vehicle's position, and environment the vehicle is interacting with
      \item Secure against varying illuminance, weather conditions, traffic, and road conditions. 
      \item Cost-adaptive based on the requirements. 
\end{itemize}
&
\begin{itemize}[noitemsep,topsep=2pt,leftmargin=*]
      \item Complicated and requires domain knowledge for implementation
      \item Cost increases exponentially with the precision of the research.
      \item Vehicle and platform-specific causing adaptability problems in large-scale implementation. 
\end{itemize}
\\
\hline
\end{tabular}
\end{table*} 
\section{TAXONOMY-BASED SURVEY}
\begin{table*}[h]
\centering

\caption{List of the publicly available datasets covered in the survey}
{\renewcommand{\arraystretch}{1.5} 
\begin{tabular}{p{2.5cm}p{5.5cm}p{1cm}p{1.5cm}p{4cm}p{1.8cm}}
\hline
\textbf{Dataset} &
  \textbf{Devices and Sensors Used} &
  \textbf{Volume} &
  \textbf{Duration} &
  \textbf{Parameter Variations} &
  \textbf{Drivers}\\ 
  \hline
  

HCRL [\citenum{MARTINELLI2020102504}] \tikz\draw[red,fill=red] (0,0) circle (.5ex);
& (Real) CANBUS sensors (51 features) &
16.7 MB &
23 hours
 & city, motorway, parking spaces &
10 \\ \hline


UAH-DriveSet [\citenum{7795584}] \tikz\draw[red,fill=red] (0,0) circle (.5ex); \tikz\draw[blue,fill=blue] (0,0) -- (0.1,0.1) -- (0.2,0) -- cycle;
& (Real) Smartphone sensors and camera, OpenStreetMap (maximum allowed speed and number of lanes) &
3.3 GB &
4 hours 20 minutes
 & traffic: motorway and secondary road;    
 driver behavior: normal, drowsy, aggressive &
6(1F, 5M) \\ \hline

SHRP2 [\citenum{shrp2-xq}]
\tikz\draw[red,fill=red] (0,0) circle (.5ex); 
\tikz\draw[blue,fill=blue] (0,0) -- (0.1,0.1) -- (0.2,0) -- cycle;
& 4 Cameras (Scene+Face+Lap+Hand), GPS, Radar, cellphone &
$~$2PB & - & varied surface condition, weather,
traffic density and lighting & $>$ 3000 \\ \hline

YawDD [\citenum{e1qm-hb90-20}] \tikz\draw[blue,fill=blue] (0,0) -- (0.1,0.1) -- (0.2,0) -- cycle;
& (Real) RGB Camera 30 Hz, Set-1 placed in front of rear mirror, Set-2 DashMounted &
4.94 GB &
2-3 hours
 & daytime, weather: sunny, cloudy, rainy &
107(50F, 57M) \\ \hline

NTHU [\citenum{Weng2016DriverDD}] \tikz\draw[blue,fill=blue] (0,0) -- (0.1,0.1) -- (0.2,0) -- cycle;
& (Simulated) IR camera   &
- &
9 hours 30 minutes
 & Varied driving scenarios, illumination conditions, drivers with and without glasses &
18(8F, 10M) \\ \hline

100-Cars [\citenum{dingus2006100}] \tikz\draw[blue,fill=blue] (0,0) -- (0.1,0.1) -- (0.2,0) -- cycle;

& (Real) Driver, Road Front - Side camera, GPS, Radar, Cellphone, Multimedia  &
6.4TB &
43,000 hours
 & varied surface conditions, weather, traffic density, and lighting &
109(43F, 66M) \\ \hline

AUC DDD [\citenum{eraqi2019driver}] \tikz\draw[blue,fill=blue] (0,0) -- (0.1,0.1) -- (0.2,0) -- cycle;
& (Real) Driver side profile images &
- & N/A & driver engaged in one of 10 tasks & V1:31(9F,22M), V2:44(15F,29M) \\ \hline

\multicolumn{6}{|l|}{
Use cases: \tikz\draw[red,fill=red] (0,0) circle (.5ex); Aggressive Driver Detection, \tikz\draw[blue,fill=blue] (0,0) -- (0.1,0.1) -- (0.2,0) -- cycle; Driver Inattention Detection, \tikz\draw[orange,fill=orange] (0,0) rectangle (0.3,0.15); Intoxicated Driver Detection

} \\ \hline

\end{tabular}
}
\label{tab:01}
\end{table*}

\subsection{Aggressive Driver Detection}


As anger is not as easily depicted on the human face and facial expressions (unlike drowsy and distracted instances), most research on transportation systems relies on vehicular information for aggressive driving assessment. Early studies in DBA relied on rule-based and fuzzy models to classify aggressive driving. \citep{7014406} in SenseFleet, a vehicle-independent driver profiling solution, used smartphone sensors, GPS, and web services to provide a driving risk score and distinguish between aggressive and calm drivers. It used the calibration phase to set up fuzzy rules and integrated contextual information of weather, time of day, and vehicular information to score the driver profile.

As the evaluation criteria for threshold-based modules are based on personal feedback, the results could be biased per individual and their preferences. Dynamic Time Warping (DTW) offers more reliable signal mapping between two-time series that is independent of individual preferences. \citep{6083078} implemented a DTW-based binary classifier that used a smartphone accelerometer to classify between normal and aggressive driving. Named MIROAD, the approach identified 97\% of the aggressive driving maneuvers accurately.

However, as the driving pattern depends on the individual driving style, traffic, and vehicle used for the experiment, pattern-matching approaches like DTW suffered from the selection of the ground truth. Recent strategies in DBA have leveraged advancements in machine learning tools and algorithms for the accurate assessment of aggressive driving behavior. \citep{7904773} employed Sequential Forward Feature Selection (SFFS) along with Random Forest to achieve an overall accuracy of 95.5\% on aggressive driver classification. However, the SFFS approach required iterative feature analysis over 78 features, which is computationally expensive and deemed inefficient in practical applications. \citep{8764567} approached the problem uniquely by representing the time-series data as words in a text document using a second-order representation of accelerometer data under the Bag of Words approach. The authors analyzed this approach over binary classification between normal and aggressive driving and multi-class classification between specific aggressive maneuvers with an accuracy of over 96\% in both cases.

Even though the machine learning approaches mentioned reported an accuracy of over 95\%, these were based on private datasets. Thus, the reported accuracy was doubtful as the data could have been collected to favor the approach. With the goal of providing an accurate assessment of driver aggressiveness in a general setting, \citep{8778416} developed a Long Short-Term Memory-Fully Convolutional Network (LSTM-FCN) model to detect aggressive behavior in a driving session using the \textit{UAH-DriveSet} \citep{7795584} dataset. All details on the public datasets discussed throughout this survey are listed in Table \ref{tab:01}. The model was able to detect aggressive behavior in a 5-minute window of a driving session with an f1-score of 0.96. Although novel and highly accurate, LSTM-FCN is computationally expensive to be implemented in an actual vehicle. Also, the 5-minute window for accurate assessment could be a life-threatening delay in the case of actual road traffic. Thus, there is a need for a simpler and computationally inexpensive approach with real-time assessment of driver behavior.

\citep{9262069} provides a data-driven risk scoring platform based on the SHRP2 \citep{shrp2-xq} dataset. The authors investigate various classical machine learning techniques and feed-forward deep neural networks to present the most accurate and computationally efficient model for risk scoring in naturalistic driving scenarios. Their investigation results in the Random Forest Classifier gave the best results, with 86\% accuracy and 0.91 f1-score. They also test and validate the performance of Random Forest in highly skewed data scenarios. Unlike other approaches focused entirely on performance, the authors in this research also discuss the practical cloud-based implementation in an actual vehicular system.

\citep{priyadharshini2022stacking} asserts that a single machine-learning algorithm is not sustainable in aggressive driver assessment. So, the authors use a stacking approach with an Artificial Bee-Colony algorithm (ABC) \citep{karaboga2010artificial} to achieve an accuracy of 98\% in driving style assessment using the HCRL \citep{MARTINELLI2020102504} dataset. The performance of the model is attributed to the feature reconstruction and the use of six-stacked base learners and an ABC meta-learner for the final prediction.

Table \ref{tab:03} summarizes all the approaches mentioned in the survey related to the detection of aggressive driving behavior.

To summarize, most of the research on the detection of aggressive driving behavior focuses on vehicular data and maneuver information to assess the risk associated with the driver. However, throughout the literature, we observe a common pattern of using a supervised approach in the detection of aggressive driving behavior. However, the data collection for aggressive driving is difficult in real-traffic scenarios, and the simulated driving experiments suffer from subjective biases when replicating aggressive driving behavior. Also, as it is a continuous behavior, unlike sudden changes as in the case of inattentive and intoxicated driving behavior, normal driving behavior for someone could be considered as aggressive for another person driving in the same scenario. Due to these reasons, the supervised approach becomes tricky when implemented in a large-scale real-world scenario. Therefore, with the limited data availability and problems in data collection and replication, semi-supervised and unsupervised driver behavior analysis need to be assessed. Also, we need to consider the practical application of the models and focus on computational efficiency and generalization along with performance and accuracy.

\begin{table*}[h]
\centering
\caption{Related Research based on Aggressive Driver Detection}
{\renewcommand{\arraystretch}{1.5} 
\begin{tabular}{p{3cm}p{5cm}p{7cm}p{1.7cm}}
\hline
\textbf{Research Ref.} &
  \textbf{Data Source} &
  \textbf{Approach} &
  \textbf{Results (Acc.)}\\ 
  \hline

  \citep{7014406} &
  Smartphone sensors + OpenStreetMap API &
  The fuzzy-logic-based risk score &
 --- \\
  \hline

     \citep{6083078} &
  Smartphone {rear-facing camera, accelerometer, gyroscope, GPS} &
  DTW + K-NN &
  97.0\%\\
  \hline

     \citep{7904773} &
 3-axis accelerometer (17 Hz) &
  GMM + SFFS + RFC &
  95.5\%\\
  \hline

       \citep{8764567} &
Smartphone accelerometer &
  PCA+BoW+ (MLP, RF, GNB, KNN) &
  0.96 (f1)\\
  \hline

     \citep{8778416} &
  UAH-DriveSet [\citenum{7795584}] & FCN-LSTM
  &
 0.96 (f1)\\
 \hline

     \citep{9262069} &
  SHRP2 [\citenum{shrp2-xq}] & kNN, SVM, RF, DNN
  &
 86\%\\
 \hline

\citep{priyadharshini2022stacking} &
  HCRL [\citenum{MARTINELLI2020102504}] & Feature Reconstruction + Base (kNN, DT, SVM, MLP, Adaboost, RF) + Meta (ABC)
  &
 98.9\%\\
 \hline

\end{tabular}
}
\label{tab:03}
\end{table*}

\subsection{Driver Inattention Detection}
Driver inattention detection research relies on driver and vehicular performance, as well as the physical and physiological reactions \citep{9618784} of the driver to the driving scenario. Inattention in driving arises when the driver is distracted or in a state of fatigue (drowsy driving). Based on this, the survey first reports on work on driver fatigue detection and proceeds on to distracted driving detection.

One of the earliest driver fatigue monitoring techniques relied on Kalman filtering and the Probabilistic Threshold approach, using the driver's gaze as the data feature to identify yawns and fatigue with more than 82\% accuracy in \citep{s121217536}. Using a Fuzzy Bayesian Network with data-fusion of vehicular, physiological, and visual (eye) features, the approach identifies driver drowsiness and alerts when the data-fusion-based metrics exceed the set threshold. The simulated experiment minimized the risk of an accident caused by the driver's drowsiness by identifying the driver's condition change from non-partial to partial sleep in approximately 0.2 seconds. However, the intrusive approach of capturing the driver's vital signals can act as a distraction, causing further harm to the driver. Also, the detection system fails if the driver wears sunglasses, has partial occlusion on their eye area, or is passing through an area with rapidly varying illuminance.

Using driving behavior and subjective drowsiness in a VR-based simulated experiment, \citep{8545427} use both classification and regression to predict driver drowsiness with 99.1\% accuracy and 0.34 Root Mean Square Error (RMSE) as early as 4.4 seconds. The system offers unobtrusive means of driver drowsiness detection, which is accurate, cost-efficient, and relies on readily available vehicular information. However, the accuracy assessment is based on the subjective Karolinska Sleepiness Scale (KSS), resulting in subjective and biased drowsiness levels for the participants. Also, the highly controlled simulated testbed doesn't reflect the actual driving scenario, putting the model's accuracy in doubt. 
 
Classical machine learning models with visual features for analysis rely on PERCLOS for driver drowsiness assessment. \citep{5648121} use the Viola-Jones face detection Algorithm \citep{viola2001rapid} with PERCLOS to detect driver drowsiness in an indoor setting with an accuracy of 94.8\%. However, the performance degrades when participants wear glasses or experience partial occlusion in the region of interest. The approach suffers as participants navigate through varying illuminance. In \citep{JO20141139}, SVM and Maximum a priori Classifier (MAP) with PERCLOS and eye closure duration (ECD) measure driver drowsiness in an IR camera-based real-driving experiment with an accuracy of 98.3\%. The IR camera-based approach resolves issues associated with wearing eyeglasses and varying illuminance and works well in low illuminance. However, the problem with partial occlusion persists. The survey \citep{Kaplan2015DriverBA} presents more on the classical machine learning and statistical methods of driver inattention detection based on visual, vehicular, and physiological signals.

Focusing on implementations based on public datasets, \citep{9892146} propose an end-to-end transformer-based DBA, TransDBC, which performs with 95.38\% accuracy on driver inattention detection based on multivariate time-series smartphone data from the \textit{UAH-DriveSet}. The research puts forth a safer and more comfortable measure for driver drowsiness using vehicle-related information from the smartphone. However, the research doesn't report the applicability of the resource-intensive transformer model in a vehicular system. Although high in accuracy, the research doesn't consider the time criticality associated with driver drowsiness assessment.
\citep{8949469} propose a multi-task CNN along with PERCLOS and Frequency of Mouth (FOM) for driver fatigue detection that achieves an accuracy of 98.72\% on the \textit{YawDD} and \textit{NTHU}. Using the Dlib face-detection algorithm \citep{king2009dlib} to first detect the face, the data is passed through a binary classifier to label based on mouth open/close and eyes open/close. This labeled data is passed through the multi-task CNN for the assessment of driver drowsiness into 3 classes: "very tired, tired, and not tired." The reported accuracy is based on a real-time system in perfect condition for the driver, lighting, and camera placement, so the proposed system still fails in scenarios with partial occlusion or varying illuminance.

In summary, most research on driver drowsiness deals with simulated data or experiments carried out in a controlled environment. Additionally, the research focuses on detection after the event rather than alert based on partial-drowsiness detection. There is still much to research on the progressive detection of driver drowsiness and efficient alert systems that are accurate in detection and computationally efficient enough to implement in real-driving scenarios.

Similarly, in the upcoming section, I will discuss research focused on driver distraction detection. \citep{QIN201923} used statistical measures such as the Tukey test, chi-square test, Nemenyi post-hoc test, and Marascuilo procedure to understand driver distraction in fatal crashes. Their study, based on fatal crash data \citep{national2015traffic} from 2010-2013, reports that inner cognitive distractions account for more than 50\% of distraction-related fatal crashes. New wearables, portable devices, and driver entertainment systems are also investigated as increasing sources of driver distraction. Based on age group, young drivers were found to have a higher probability of distraction, and based on sex group, young females seem to have a higher probability of engaging in in-vehicle distraction-related activities. This research paved the way for upcoming researchers working on understanding, analyzing, and detecting driver distraction.

Prioritizing the work done on detecting cognitive distractions, one of the earliest and most prominent research in driver distraction detection comes from \citep{1336359}. Using a priori threshold for detecting the pupil in IR images of the driver and using Kalman filtering to keep a continuous track of the detected pupil across multiple image frames, the approach predicts the driver's state using a Finite State Machine. This, along with the pupil features passed into a fuzzy interface, processes the driver's vigilance. However, as with any fuzzy system, the thresholds set are subjective to the research, participants, and other constraints, and there is no standardized evaluation metric for the actual assessment of the model.

In the same year, \citep{1398976} researched standardized means of driver workload assessment. Using a simulated test bed, the research relied on a decision tree for workload assessment under varying conditions of features with 81\% accuracy while using all the features and 60\% when removing the eye and gaze features and just using the vehicular features for assessment. This standardized metric-based workload reporting and use of machine learning paved the way for more research in the field of driver distraction detection.

More recent advancements in research on driver distraction detection come from the implementation of deep learning approaches. \citep{8678436} used a front-facing camera to record and detect the driver performing one of the seven assigned actions, of which three reflected distracted driving: using an in-vehicle radio/video device, answering the mobile phone, and texting. Using Gaussian Mixture Modelling (GMM) to detect drivers from the video stream and AlexNet to perform behavioral classification, the module achieved an accuracy of 81.6\%. The classification for each image is carried out in about 60 seconds, providing a real-time solution for the time-critical assessment of driver distraction.

However, the research discussed so far is based on a private dataset with data collection and feature engineering fixed to benefit the proposed approach. \citep{eraqi2019driver} introduced a comprehensive public dataset for driver distraction assessment, the AUC Dataset. The dataset comprises side profile images of drivers engaged in distractive activities. From the dataset, raw images, skin-segmented images (using Multivariate Gaussian Naive Bayes classifier (MG-NBC)), face images, hands images, and “face+hands” images were used to train an ensemble of convolutional neural networks with 90\% classification accuracy.

Analysis based on the \textit{AUC} dataset relied on the driver's side profile images. The discontinuous image sequence meant that there was no means of progressive distraction identification. The dataset also did not consider the vehicular features and the external environment the driver interacts with.

\citep{dingus2006100} published the \textit{100-Car} dataset, a first-of-a-kind collection of naturalistic driving data from novice and experienced drivers, including scenarios such as crashes and near-crashes. \citep{doi:10.1056/NEJMsa1204142} examined the correlation between distracted driving and accident scenarios. Based on multi-model logistic regression analysis, the authors reported that two activities, using a cell phone for dialing and reaching out to grab something apart from the cell phone, had the highest correlation with the likelihood of an accident. This seems extremely prevalent for novice drivers who represent the majority of traffic in day-to-day driving.

The Strategic Highway Research Program 2 (SHRP2) \citep{shrp2-xq} aimed to provide a better assessment of driver distraction in the wild and presented a dataset with a 4-camera view of the driving experiment, along with GPS and cellphone features. \citep{liu2021identification} used the \textit{SHRP2} dataset in their assessment of distracted driving. Using the vehicular kinematics from the dataset, the researchers compared the performance of LSTM-NN, SVM, and AdaBoost, with LSTM-NN performing the best because of its ability to keep track of long-dependent time series data. Gradient-boosted decision tree recursive feature elimination (GBDT-RFE) and random forest recursive feature elimination (RF-RFE) were used to evaluate feature importance, and using LSTM-NN with the identified important features, the system detected distracted driving with an accuracy of 88\%, providing an accurate assessment of driver distraction in naturalistic driving scenarios. A similar approach was employed in \citep{WANG2022103561} with the same dataset. Instead of an LSTM-NN, a Bi-LSTM network with an additional attention mechanism was employed, which improved the accuracy of the system to 91.226\%. The latter also promised a real-time assessment of driver distraction with a runtime of 19.34 ms. 

However, the current dataset available still fails to capture all driver distraction events due to the risks associated with data collection. Therefore, there is a need for a shift in research towards semi-supervised and unsupervised methods of driver distraction identification that are robust, real-time, and computationally efficient for everyday driving situations. Table \ref{tab:04} summarizes all the approaches discussed in this survey for inattentive driver detection.

\begin{table*}[h]
\centering
\caption{Related Research based on Driver Inattention Detection}
\renewcommand{\arraystretch}{1.5} 
\begin{tabular}{p{0.5cm}|p{4cm}p{5cm}p{6cm}p{1.2cm}}
        \hline
         & \textbf{Research Ref.} & \textbf{Data Source }& \textbf{Approach} & \textbf{Results}\\
        \hline
         \multirow{6}{*}{{\rotatebox[origin=r]{90}{Drowsy Driver Detection}}}
        & \citep{s121217536} & (Sim.) Smartphone + Physiological sensors & Fuzzy-Bayesian Network & 82\%  \\
        & \citep{8545427} & (Sim.) VR + KSS + Sim. vehicle data & Multiple Linear Regression (MLR), Decision Tree Classifier & 99.1\% \\   
        & \citep{5648121} & (Sim.) Driver facing Webcam & Viola Jones Algorithm + PERCLOS + Threshold based classifier & 94.8\% \\
        & \citep{JO20141139}  & (Real) Driver facing IR Camera & Adaboost + PCA + SVM, MAP  & 98.3\% \\
        & \citep{9892146}  & (Real) UAH-Driveset [\citenum{7795584}] & TransDBC (end-to-end transformer)  & 95.38\% \\
        & \citep{8949469}  & (Real) YawDD [\citenum{e1qm-hb90-20}] + NTHU [\citenum{Weng2016DriverDD}] & Multitask CNN + PERCLOS + FOM  & 98.72\% \\
        \cline{1-5}
        \cline{1-5}
        \hline
        \multirow{7}{*}{{\rotatebox[origin=r]{90}{Distracted Driver Detection}}}
        & \citep{s121217536} & (Real) FARS (NHTSA) [\citenum{national2015traffic}] & Tukey test, chi-square test, Nemenyi post-hoc test, Marascuilo procedure & - \\
        & \citep{1336359} & (Real) IR-Camera & Kalman Filtering + FSM + Fuzzy & -\\
        
        & \citep{1398976} & (Sim.) Monocular gaze-tracker + Head tracker + Sim. CANBUS  & ANOVA + Decision-tree & 81\% \\
        & \citep{8678436}  & (Real) Driver facing RGB Camera & GMM + AlexNet  & 81.6\% \\
        
        & \citep{doi:10.1056/NEJMsa1204142}  & (Real) 100-Cars [\citenum{dingus2006100}] & Multimodel Logistic Regression  & - \\
        
        & \citep{eraqi2019driver}  & (Real) AUC [\citenum{eraqi2019driver}] & MG-NBC + AlexNet, ImageNet  & 90\% \\  
        & \citep{liu2021identification}  & (Real) SHRP2 [\citenum{shrp2-xq}]  & GBDT-RFE, RF-RFE + LSTM  & 88\% \\
         & \citep{WANG2022103561}  &  & Bi-LSTM (with Attention)  & 91.23\% \\
         \hline
    \end{tabular}
    \label{tab:04}
\end{table*}

\subsection{Intoxicated Driver Detection}
\begin{table*}[h]
\centering
\caption{Related Research based on Intoxicated Driver Detection}

{\renewcommand{\arraystretch}{1.5} 
\begin{tabular}{p{4cm}p{5cm}p{6cm}p{1.7cm}}
\hline
\textbf{Research Ref.} &
  \textbf{Data Source} &
  \textbf{Approach} &
  \textbf{Results (Auc.)}\\ 
  \hline

   \citep{verster2014effects} &
  Sim. Vehicle feature &
  SDLP + ANOVA &
  ----\\
  \hline

     \citep{LI201561} &
  Sim. Vehicle feature &
 Bottom-up signal segmentation + SVM &
  80\%\\
  \hline

       \citep{9517124} &
  Driver-facing video camera &
 VCG + Dense-Net &
  89.62\%\\
  \hline

  \citep{9258992} &
  Alcohol + Temperature Sensor, Camera, &
  SVM, KNN, Bayes Classifier, Neural Network &
 97\% \\
  \hline
                           
\end{tabular}
}
\label{tab:05}
\end{table*}

Research on intoxicated driver detection mostly relies on psychological and biomedical sensors that provide an accurate assessment of driver intoxication. However, these sensors also pose a threat of further casualty with their intrusive nature and require specific modes of operation, placement, and orientation. There are still few types of research that investigate the detection of driving under influence using visual and vehicular features.

Using a simulated highway scenario, \citep{verster2014effects} examined the effect of alcohol hangovers on driving performance. The weaving of the car, expressed as the standard deviation of the lateral position (SDLP), was the main parameter of interest to establish the effect of alcohol hangovers. The level of hangover was established using statistical analysis of data collected on control day (without alcohol) and hangover day. Statistical analysis of vehicular and subjective information revealed that alcohol hangovers significantly impacted driving performance with an increase in vehicle weaving and a lapse in attention. This research is one of a kind in establishing driver behavior while driving hungover and the effects it produces on the vehicular trajectory. However, the subjective assessment of intoxication is biased by the participants' self-reporting. The research also focuses on establishing the correlation rather than detecting and alerting the driver under intoxication.

In \citep{LI201561}, an accuracy of 80\% was achieved in differentiating between normal and drunk driving in a simulated highway driving experiment. Using multi-variant time-series data of vehicle lateral position and steering angle, the time series are segmented using a bottom-up segmentation algorithm and then fed into an SVM classifier. The system is able to identify drunk driving in 25 seconds of driving performance, providing real-time assessment and potential alert systems. However, the experiment runs on a small test bed and focuses on the binary classification of the drunk state. Additional research on large sample sizes and traffic conditions with risk-assessment-based classification is required in the practical application of drunk driver detection.

In \citep{9517124}, drunk drivers are identified using a two-stage deep neural network. The driver's age is estimated using a VCG network in the first phase, and Dense-Net is used to detect the facial features of the driver in the second stage. Using these two networks, the research claims to identify drunk drivers with an overall accuracy of 89.62\%. Though novel in its approach, the research is entirely based on an indoor setting and video recording in perfect illuminance condition. As mentioned before, the facial features-based approach requires the camera to be set up in a specific orientation and under specific illuminance conditions, which may not be applicable in a practical application. This research also doesn't mention the model assessment time, which is critical to avoid any casualties caused by drivers under the influence.

\citep{9258992} addresses the non-intrusive detection of intoxicated drivers using the sensor fusion approach. Three sensors -- an alcohol in the environment sensor, a temperature sensor, and a video camera for pupil diameter assessment -- are fused and passed into multiple machine-learning models. The system is able to identify intoxicated drivers with 97\% accuracy using a neural network. The research also identifies the gas sensor (to detect alcohol in the environment) and the pupil to be the most relevant features for intoxicated driver detection. However, this approach still relies on the proper placement of sensors inside the vehicle and the driver to operate the vehicle in a specific pattern, which is not possible for real-time application in intoxicated driving. The system suffers from the partial opening of car windows, changes in illuminance while driving, and person-specific psychological changes while intoxicated. Table \ref{tab:05} summarizes all the approaches discussed for intoxicated driver detection in this survey.

In summary, intoxicated drivers contribute to more than 20\% crashes. Compared with other behavioral assessment techniques, there appears to be limited research and resources in identifying intoxicated drivers, which could be attributed to the risk associated with the experiment and the repeatability of research while driving under the influence. Studies that determine the subjective tolerance of individuals to alcoholic and non-alcoholic drinks are also missing in the context of detecting intoxicated drivers. With the legalization of medicinal cannabis and the traditional methods of detection failing on recreational drugs, there is an immediate need for comprehensive real-time detection of drivers under the influence of drugs. However, thorough research is required to determine the learning and analyzing parameters, considering the variability of drugs and the subjective tolerance of individuals, and to ensure that the learning also works in such scenarios for the better implementation of intoxicated driver detection.

\section{DISCUSSION AND OPEN PROBLEMS}
Visual and non-visual features have proven to be efficient in analyzing driver behaviors across multiple modalities. Initially starting as a threshold-based assessment, the analysis modules have evolved into complicated and highly accurate deep-learning modules. However, there are still a few open-ended problems that need to be resolved for a robust solution.

One of the major concerns associated with the analysis is the lack of data and the limitations of data collection in risky driving scenarios. Both private and public datasets collected using real-world driving have limitations with experiments for inattentive, intoxicated, and aggressive driving. Real-world experiments suffer from variability in the road, weather, and illuminance conditions. Some papers discussed in the survey resolve this issue with simulated test beds, which give researchers freedom over the variability, repeatability, and scalability of the experiment. However, the question now arises on whether the results obtained from the simulated test bed can be generalized over real-world traffic conditions. I am currently working on transfer learning based on a multi-platform experimental setup to address this issue, where data from both simulated and real-world experiments are analyzed to form a generalizable solution for assessing complicated driving behavior.

Sensor fusion demonstrates the most promising results for DBA. The system benefits from limitations associated with individual sensors. However, researchers \citep{NASRAZADANI2022105} and \citep{doi:10.1056/NEJMsa1204142} are limited to using only one of the features presented from multi-feature models, while \citep{9258992} suffer from computational efficiency issues while trying to combine multiple modalities for Driver Behavior assessment. Therefore, the current issue that needs addressing is the research on efficient sensor fusion, feature processing, and extraction techniques for the scalable applicability of Driver Behavioral Analysis.

The use of DBA in applications like Insurance Pricing \citep{he2018profiling} raises concerns about privacy and security, especially in visual features-based DBA. While these features offer higher accuracy, there are instances where the system might mislabel any driver based solely on their facial expression, raising concerns about biases in the implementation of these applications. Thus, there is a need for research in developing algorithms and modules that protect the privacy and security of the individual while maintaining the efficiency and accuracy of the assessment.

Recently, we have seen increasing interest in connected vehicular perception and information sharing among intelligent vehicles. Even though most of the papers mentioned in the summary investigate driver behavior at an individual level, there is a need to address the behavioral change as the driver interacts with vehicles, pedestrians, and other nodes in the traffic. Following this discussion, \citep{app112110462} implemented an information-sharing Vehicle-to-Vehicle (V2V) communication using 5G, Road Side Units (RSUs) as nodes, and vehicles as edge devices for accident prevention. However, one must consider the computational overload associated with the number of vehicles. The research also illustrates the exponential rise in false alarm rates as the number of vehicles in the network increases. Another issue associated with connected vehicles is the security and standardization of information sharing. Therefore, the security and computational load associated with V2V information remain an open problem that needs to be resolved for the efficient implementation of DBA in the future.

\section{CONCLUSION}
In this survey, I have discussed literature focused on using visual and vehicular features for efficient driver behavior analysis. I have explored the progression of research with advancements in sensor and computational capabilities. The integration of visual and vehicular information can provide a more accurate and effective analysis of driver behavior, leading to the development of better safety measures and reduced traffic accidents. However, there are still open problems that limit the applicability of these approaches. I hope this survey serves as a guideline for future research directions in driver behavior analysis.

\bibliography{conference_101719}

\end{document}